\documentclass{article}
\usepackage{spconf,amsmath,graphicx}
\usepackage{multirow}
\usepackage{booktabs} 
\usepackage{graphicx}
\usepackage{subfigure} 

\title{Type I attack for generative models}

\name{Chengjin~Sun$^{\star}$ \qquad Sizhe~Chen$^{\star}$ \qquad Jia Cai$^{\dagger}$ \qquad Xiaolin~Huang$^{\star}$}
\address{$^{\star}$Department of Automation, Shanghai Jiao Tong University, Shanghai, P.R. China. \\
	$^{\dagger}$HuaWei Technologies Co.,Ltd, Hangzhou, P.R. China.}

\begin{document}
%
\maketitle
\begin{abstract}

Generative models are popular tools with a wide range of applications. Nevertheless, it is as vulnerable to adversarial samples as classifiers. The existing attack methods mainly focus on generating adversarial examples by adding imperceptible perturbations to input, which leads to wrong result. However, we focus on another aspect of attack, i.e., cheating models by significant changes. The former induces Type II error and the latter causes Type I error. In this paper, we propose Type I attack to generative models such as VAE and GAN. One example given in VAE is that we can change an original image significantly to a meaningless one but their reconstruction results are similar. To implement the Type I attack, we destroy the original one by increasing the distance in input space while keeping the output similar because different inputs may correspond to similar features for the property of deep neural network.  Experimental results show that our attack method is effective to generate Type I adversarial examples for generative models on large-scale image datasets.

\end{abstract}
\section{Introduction}
\label{sec:intro}

Generative models are considered to be one of the greatest inventions in the field of AI. Two most representative types are: the generative adversarial networks (GAN) \cite{goodfellow2014generative} and the variational autoencoder (VAE)  \cite{kingma2013auto}.
They have many applications, such as auto-programming \cite{mou2015end}, compressing information \cite{gregor2016towards}, interactive image editing \cite{zhu2016generative,dekel2018sparse}, sketch2image \cite{chen2018sketchygan,sangkloy2017scribbler}, and other image-to-image translation tasks \cite{zhu2017unpaired,wang2018high}.

It is now well-known that DNNs are vulnerable to adversarial attacks. In \cite{gondim2018adversarial}, it has been found that imperceptive perturbations on the input of autoencoder cause the reconstruction result to change significantly.  In statistical, this attack corresponds to Type II attack on classifiers, i.e., manipulating the input by adding imperceptible perturbations \cite{szegedy2013intriguing,madry2017towards,moosavi2016deepfool,carlini2017towards} or changing the semantic attributes of images \cite{shamsabadi2019colorfool,hosseini2018semantic,joshi2019semantic},  which has attracted many attention of researchers and becomes a big concern. Very recently, we find that it is also possible to implement Type I adversarial attack \cite{tang2019adversarial}. Type I attacks on generative models, i.e., changing the input significantly but leading to similar output, is potentially as dangerous as attacks on classifiers and also meaningful to investigate. For instance, Type I attack could do harm to the information transition because autoencoders are widely used for compressing information. A malicious noise image which is far from the clean one may lead to a reconstruction output which is similar to the origin. Type I adversarial examples can also point out the weakness of the generative models and are valuable for enhancing the robustness of the network.

\begin{figure}[t]
	\centering
	\includegraphics[width=2.7in]{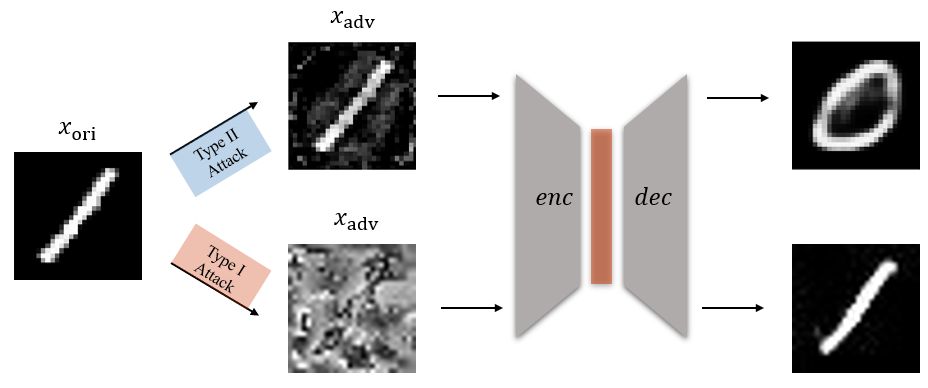}
	\caption{Type I and Type II adversarial attacks on VAE. }
	\label{typeI_typeII}
\end{figure}

In this paper, we design Type I attack on generative models. The difference between Type I and Type II attacks could be understood by the following example. As illustrated in Fig \ref{typeI_typeII}, the top is the Type II attack where we slightly disturb ``1'' such that the adversarial digit is still ``1'' but its reconstruction result is another digit ``0''. Type I attack is shown at the bottom. Although the noisy and meaningless image is fed to the VAE,  the generative model outputs a clean example ``1''. 

Mathematically, the Type I attack, i.e., the input is changed significantly but the generator still gets a similar output, is defined as the following,

\begin{equation}
	\begin{aligned}
	\mathrm{From} \quad  x \quad \ &\mathrm{Generate}\quad x^{\prime}= \mathcal A(x)\\
	s.t. \quad\ &G(x^{\prime})=G(x)\\
	&\left \|x-x^{\prime}\right \|>\epsilon,
	\end{aligned}
\end{equation}
where $x$ is the input and $x^{\prime}$ represents the adversarial example. $G$ denotes the generative models and $\epsilon$ is the threshold.

The underlying reasons for Type I attack and Type II attack are different on features as explained in \cite{tang2019adversarial}.  The existence of Type II adversarial examples is due to the unnecessary feature considered by the usual generative model but not used by an ideal one. So the variant in the unnecessary feature ignored by the oracle makes the output of the usual generative model change greatly. Conversely, the missing feature is taken into account in the ideal one but is omitted by the usual one.  So attacking missing feature causes Type I attack. The essential difference between both attacks makes the defense designed for Type II do not help the improvement of  Type I robustness. Thus, Type I attack should be simultaneously considered together with Type II to strengthen the robustness of generative models. 

In this paper, we perform Type I attack on two most representative generative models: VAE and StyleGAN.  For VAE, the adversarial image is generated by increasing its distance to the clean one and keeping their outputs similar. Another way to attack is by updating the latent variable which recovers the adversarial image through the decoder. Benefiting from the restriction in the latent space and the gradient related to increasing the distance between the adversarial example and the original, the attack is achieved. The datasets used to perform Type I attack on VAE are  MNIST, SVHN, and CelebA. For StyleGAN,  the attack is implemented by updating the intermediate latent space which is directly related to the generated images' styles.

The rest of this paper is organized as follows. In Section II, we introduce the techniques of Type I adversarial attack. Section III evaluates the proposed attack on VAE and StyleGAN. In Section IV, a conclusion is given to end this paper.

\section{Type I attack on Generative Models}
\label{sec:pagestyle}

In this paper, we focus on the most representative generative models: VAE and StyleGAN. VAEs are neural networks consisting of an encoder $e(\cdot)$ and a decoder $d(\cdot)$. The encoder outputs the parameters of the latent distribution from the input, and then the decoder samples the latent distribution and reconstructs something similar to the input. StyleGAN \cite{karras2019style} is a representative GAN, which has a clear hierarchy of features, can generate ultra-high-resolution samples. StyleGAN is composed of two sub-networks: a non-linear mapping network $f: \mathcal{Z}\rightarrow\mathcal{W}$ which maps the latent code $z$ to an intermediate latent code $w=f(z)$, and a synthesis network $Gs(\cdot)$ which starts from a constant, and receives styles from $w$ after affine transformations to control adaptive instance normalization every time before upsampling image. Therefore, the generator of StyleGAN can be represented as $G=Gs\circ f(z)$.

For Type I attack on VAE, it is required to generate an adversarial image which is totally different from the origin. Here, we propose the attack on the image space to generate random noise and keep their reconstruction outputs similar. We also could attack the latent space. That means the gradients do not merely propagate to the image space, but further to the latent space. When attacking the image space, we push the adversarial image away from the original one while minimizing the distance between their reconstruction outputs.

Mathematically, the above idea can be described as the following function,

\begin{eqnarray}\label{vae-loss_x}
L_{x}=\left \|d(e(x))-x_{\rm ori}\right \| +\lambda*\left \|x-x_{\rm ori}\right \|,
\end{eqnarray}
where $x_{\rm ori}$ denotes the original image and $x$ is the input variable to optimize. The first part of the loss function makes sure the similarity of the outputs while the second part destroys the input to a meaningless one. Hyper-parameter $\lambda$ aims to balance the two parts.
Notice that the norm here could be replaced by many distances. In this paper, we use $l_1$-norm distance when attacking SVHN and CelebA and $l_2$-norm distance for MNIST.

Also, we can find the adversarial example by searching in the controllable latent space and then decoding it so that the adversarial examples are expected to follow a known distribution. In this way, we can capture more features instead of generating random sharp noise. In the process of attacking the latent space, we enlarge the distance within a specific threshold in the latent space and keep the similarity of the reconstruction outputs simultaneously, as follows,
\begin{equation}\label{vae-loss_z}
\begin{aligned}
L_{z}=\left \|d(e(d(z)))-d(e(x_{\rm ori}))\right \| +\\
\lambda*ReLU(\varepsilon-\left \|z-e(x_{\rm ori}) \right \|),
\end{aligned}
\end{equation}
where $z$ is the latent variable which we need to optimize, $d(z)$ is the adversarial image and $\varepsilon$ is a threshold which restricts $z$ on the mainfold of the latent space.

For StyleGAN, instead of revising the latent variable of the mapping network, we choose to optimize in the intermediate latent space because it is disentangled and directly controls the feature of the generation output through learned affine transformations. Therefore, we minimize the following objective function to make the disentangled intermediate latent variable change significantly but the output still similar to the origin, 

\begin{equation}\label{stylegan-loss}
\begin{aligned}
L_{s}=\frac{1}{n_1}\sum\left \|Gs(w)-Gs(w_{\rm ori})\right \| +\\
\lambda*ReLU(\varepsilon-\frac{1}{n_2}\sum\left \|w-w_{\rm ori}\right \|),
\end{aligned}
\end{equation}
where $Gs(w)$ and $Gs(w_{\rm ori})$ are generated images of StyleGAN which correspond to the adversarial intermediate latent vector $w$ and original vector $w_{\rm ori}$ respectively. $n_1$ and $n_2$ are the size of the generated image and the feature vector. Here we choose $l_1$-norm to optimize for the reason that restriction in every pixel contributes to generating clearer images with more details rather than a fuzzy one.

To show whether the style feature vector changes greatly, we use the following criteria to measure the deviation:
\begin{eqnarray}\label{Dev}
Dev=\frac{1}{n}\left \|\frac{w-w_{\rm ori}}{w_{\rm ori}} \right \|_2*100 \%  ,
\end{eqnarray}
where $n$ is the size of $w_{\rm ori}$.

In Eq. \eqref{vae-loss_x}, \eqref{vae-loss_z} and \eqref{stylegan-loss}, $\lambda$ reflects the balance between the variation in the input and the similarity in the output. In our method, $\lambda$ is set to a constant when attacking VAE. For StyleGAN, $\lambda$ varies for different iterations. At the start of the optimization, we set a large $\lambda$ to enlarge the change of the input and allow the generated image changing to a totally different one with different features. After that, $\lambda$ decreases to pull the generated image back so that it is still as same as the original one. Specifically, inspired by \cite{tang2019adversarial}, a self-adaptive weight strategy is designed for
$\lambda$ to maintain such equilibrium:
\begin{equation}\label{weight strategy}
\begin{aligned}
\lambda_{k+1}=&\lambda_{k}+\alpha(\beta \frac{1}{n_1}\sum\left \|Gs(w)-Gs(w_{\rm ori})\right \|\\
&-ReLU(\varepsilon-\frac{1}{n_2}\sum\left \|w-w_{\rm ori}\right \|))\\
&+min\{ReLU(\varepsilon-\frac{1}{n_2}\sum\left \|w-w_{\rm ori}\right \|)-\widehat{L}_{w},0\},
\end{aligned}
\end{equation}
where $\beta$ controls the balance between $\frac{1}{n_1}\sum \Vert Gs(w)-Gs(w_{\rm ori})\Vert $ and $ReLU(\varepsilon-\frac{1}{n_2}\sum\left \|w-w_{\rm ori}\right \|)$ in the iteration process. At the begining, it is set as follows,

\begin{equation}\label{beta}
\begin{aligned}
\beta=\frac{ReLU(\varepsilon-\frac{1}{n_2}\sum\left \|w-w_{\rm ori}\right \|)}{\frac{1}{n_1}\sum\left \|Gs(w)-Gs(w_{\rm ori})\right \|},
\end{aligned}
\end{equation}
where $\widehat{L}_{w}$ is a loss threshold related to the feature space and this loss term concentrates more on the similarity of the generated images.
A larger $\beta$ means larger diversity in features while a smaller $\beta$ makes the generated image corresponding to the adversarial input still seem like the origin.

The adversarial samples are generated in an iterative process by minimizing loss function $L$. The update could be generally described as follows.

Attack on the image space of VAE case:
\begin{eqnarray}\label{updatex}
x^{k+1}=x^{k}-\eta \frac{\partial L_{x}(x)}{\partial x}\big|_{x=x^{k}}.
\end{eqnarray}

Attack on the latent space of VAE case:
\begin{eqnarray}\label{updatez}
z^{k+1}=z^{k}-\eta \frac{\partial L_z(z)}{\partial z}\big|_{z=z^{k}}.
\end{eqnarray}

Attack on the intermediate latent space of StyleGan case:
\begin{eqnarray}\label{updatew}
w^{k+1}=w^{k}-\eta \frac{\partial L_s(w)}{\partial w}\big|_{w=w^{k}}.
\end{eqnarray}

Specifically, we set $x^0=x_{\rm ori}$, $z^0=e(x_{\rm ori})$ and $w^0=w_{\rm ori}$. By iterative update, we gradually change the variable($x$, $z$, $w$) until the attack is successful, i.e., the distance between the adversarial input and origin of both generative models reaches a specified threshold $\zeta$ and the difference of their outputs can not exceed a threshold $\xi$.
 
\begin{figure}[htb]
	\centering
	\subfigure[]{
	\begin{minipage}[t]{1\linewidth}
			\centering
	\includegraphics[width=\hsize]{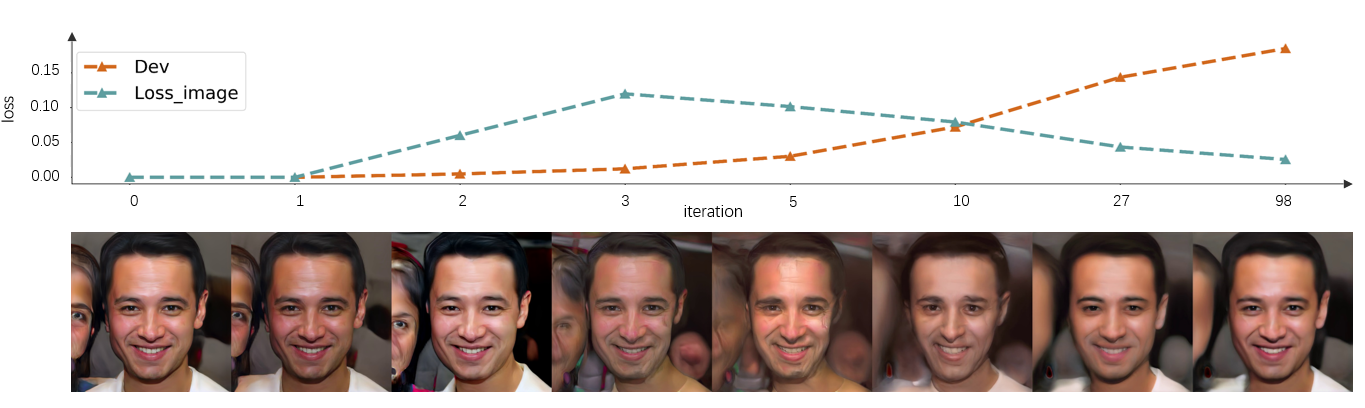}

	\label{f1}
    \end{minipage}
    }
    \subfigure[]{
    	\begin{minipage}[t]{1\linewidth}
    		\centering
	\includegraphics[width=0.99\hsize]{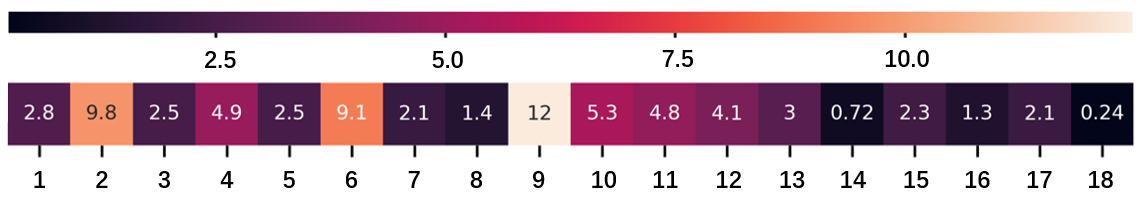}
	\label{f2}

\end{minipage}
    }
\centering
\caption{(a) Type I attack on StyleGAN. The top line shows the deviation of the feature and the loss in the output space. Images of different attack epoch ($0, 1, 2, 3, 5, 10, 27, 98$ from left to right) are displayed at the bottom. (b) Rate of the change in each dimension of the 18-dimensional feature vector.
}
	\label{loss_change}
\end{figure}

Fig. \ref{f1} shows the Type I attack process on the StyleGAN. With the increasing of iterations, the deviation in the feature space is increasing. The generated image of the adversarial feature vector first gets pushed away from the original one and becomes a totally different face. Then it is pulled back to be similar to the origin because of the training strategy.  Fig. \ref{f2} illustrates the adversarial feature changes a lot in every dimension.

\section{Experiments and Results}
\label{sec:typestyle}

In this section, we validate the proposed Type I attack on VAE and GAN  to show how the method can change the input significantly but keep the output of the attacked generator unchanged. The datasets we used are MNIST, SVHN, and CelebA. The structure of VAE is modified on \cite{gondim2018adversarial, tang2019adversarial}. The GAN is the StyleGAN pretrained in FFHQ. The details about the models, attack implementation and attack results are given in the supplemental materials.

\subsection{Type I attack on VAE}
\label{sssec:subsubhead}

First, we use the proposed method to attack VAE on MNIST, SVHN, and CelebA. The reconstruction errors of MNIST, SVHN, and CelebA are 0.099, 0.036, 0.040 measured by the root means squared deviation. The attack target is to disturb the input significantly to generate a meaningless image but its reconstruction output is similar to the origin.

\begin{figure}[!t]
	\centering
	\includegraphics[width=3.0in]{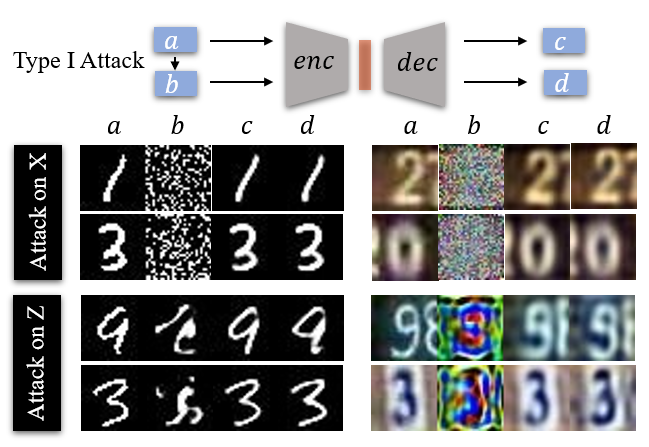}
	\caption{Type I attack on MNIST and SVHN. The attack process is shown on the top: a. the original image; b. Type I adversarial example; c. reconstruction result of original; d. reconstruction result of adversarial example.
		The left column indicates Type I attack methods. Attack on MNIST shows on the left and attack on SVHN is shown on the right.
	}
	\label{mnist_svhn_typeI}
\end{figure}

In Fig. \ref{mnist_svhn_typeI}, we show some typical adversarial examples of Type I attack for VAE  on MNIST and SVHN respectively. More examples are provided in supplemental materials. Each image pair $(x_{\rm ori},x_{\rm adv})$ satisfies: the original input $x_{\rm ori}$ and the adversarial example $x_{\rm adv}$ are totally different but the reconstruction results of them are similar. Results are illustrated in  Table \ref{attack_ms}  quantitatively. The distance we used to measure is the root mean squared deviation in each pixel (which is normalized to $[0, 1]$). We can see that although the adversarial images are different from the origin, their outputs are still similar. Utilizing the threshold criterion defined before, all the Type I attack are evaluated to be successful. 

\begin{table}[!t]
	\caption{The average distance of the adversarial examples and original ones both in the input and reconstruction output.	
	}
\scalebox{0.88}{
	\centering
	\begin{tabular}{lllll}
		\toprule
		& \multicolumn{2}{c}{Attack on X}&\multicolumn{2}{c}{Attack on Z} \\
		\midrule
		&Dis\_input      & Dis\_output       & Dis\_input & Dis\_output \\
		\midrule
		MNIST     &0.611 & 0.059         &0.213 &0.095 \\
		SVHN     &0.270       &0.021      &0.247  &0.042  \\
		\bottomrule
	\end{tabular}}
	\label{attack_ms}
\end{table}

\begin{figure}[t]
	\centering
	\includegraphics[width=2.7in]{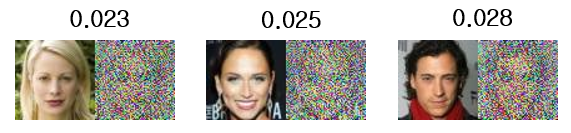}
	\caption{Adversarial example pairs with their reconstruction distance measured by the root means squared deviation.}
	\label{celeba_typeI}
\end{figure}

For CelebA, some adversarial examples are given in Fig. \ref{celeba_typeI}.  In each pair, the left is the original image, and its Type I adversarial example shows on the right.  Note that all the distances above images are below $0.1$. Accordingly, the reconstruction results of each pair can be recognized as the same person. The average attack performance is shown in Table \ref{CelebA}. Besides the root mean square error, we use the most popular face recognizers FaceNet \cite{schroff2015facenet} and Insightface \cite{deng2019arcface} to see whether the two reconstruction results corresponding to the adversarial image and original are the same person. For FaceNet, two people can be recognized as the same one only if their distance is below 1.2. In Insightface, when the similarity of two images exceeds 0.6, they could be identified as the same person. We can get the conclusion from Table \ref{CelebA} that all the attack is successful.

\begin{table}[t]
	\caption{The average difference of the adversarial examples and original ones measured both in the input and reconstruction output when performing attack on the image space. 
	}
	\centering
	\scalebox{0.88}{
	\begin{tabular}{lllll}
		\toprule
		&Dis\_input &Pixel distance &FaceNet &Insightface\\
		\midrule
		CelebA   &0.308  &0.029  & 0.453 &0.878        \\
		\bottomrule
	\end{tabular}}
	\label{CelebA}
\end{table}

\subsection{Type I attack on StyleGAN}
\begin{figure}[!t]
	\centering
	\includegraphics[width=3.0in]{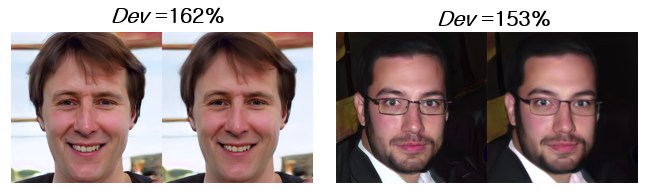}
	\caption{Type I attack on StyleGAN. The left image
		is the generated face corresponding to the original feature vector. The right image is the attack result corresponding to the adversarial one. The number on the top  is the deviation defined before.}
	\label{stylegan_typeI}
\end{figure}

Next, we evaluate the proposed attack method on StyleGAN. In Fig. \ref{stylegan_typeI},  we display some examples of the Type I attack on StyleGAN. The attack result shows a conflict. When the distance of feature vector $w_{\rm adv}$ and $w_{\rm ori}$  is large, it is expected the corresponding outputs are totally different persons with different styles. However, under the attack, the deviations are all above 150\%, the generated faces are still similar. The attack performance can be found in Table \ref{stygan}, showing that although features vary a lot, the generated image is still be identified as the same one.

\begin{table}[!t]
	\caption{The average difference between the adversarial intermediate latent vector and original one and their corresponding generated images.
	}
	\centering
	\scalebox{0.88}{
	\begin{tabular}{lllll}
		\toprule
		&Dev &Pixel distance &FaceNet &Insightface\\
		\midrule
		StyleGAN    &0.188 & 0.049 & 0.332 &0.923        \\
		\bottomrule
	\end{tabular}}
	\label{stygan}
\end{table}
\section{CONCLUSION}
\label{sec:print}
In this paper, we propose Type I attack for the generative models which aims at generating a totally different and meaningless adversarial example whose output is similar to the original. Specifically, we design Type I attack on VAE and StyleGAN and the experiments show that the proposed method successfully generates Type I adversarial examples to cheat generative models. Except for the Type II attack, Type I attack is also important to understand the generative models and worth researching because the underlying mechanisms of them are different. Type I attack for generative models can also be used to evaluate model performance and promote progress in defense methods.

\bibliographystyle{IEEEbib}
\bibliography{refs}

\end{document}